\documentclass{article}

\PassOptionsToPackage{numbers, compress}{natbib}
\usepackage[nonatbib,preprint]{neurips_2021}

\usepackage{graphicx}
\usepackage[caption=false]{subfig}
\usepackage{natbib}
\usepackage{amssymb,amsmath,amsthm,bm}
\usepackage{graphicx}
\usepackage{epstopdf}
\usepackage[utf8]{inputenc} 
\usepackage[T1]{fontenc}    
\usepackage{hyperref}       
\usepackage{url}            
\usepackage{booktabs}       
\usepackage{amsfonts}       
\usepackage{nicefrac}       
\usepackage{microtype}      
\usepackage{xcolor}         
\usepackage{verbatim}
\usepackage{upgreek}

\newtheorem{definition}{Definition}
\newtheorem{proposition}{Proposition}
\newtheorem{lemma}{Lemma}
\newtheorem{remark}{Remark}

\newcommand{\hw}[1]{{{#1}}}
\newcommand{\old}[1]{{\iffalse{#1}\fi}}

\newcommand{\R}{{\mathbb{R}}}
\newcommand{\E}{{\mathrm{E}}}

\title{Ensemble Defense with Data Diversity:\\ Weak Correlation Implies Strong Robustness}

\author{%
  Renjue Li$^{1,2}$, Hanwei Zhang$^{3,4}$, Pengfei Yang$^{1,2}$, Cheng-Chao Huang$^{5,6}$,\\
  {\bf Aimin Zhou$^4$, Bai Xue$^{1,2}$, Lijun Zhang$^{1,2}$}\\
  $^1$ State Key Laboratory of Computer Science, ISCAS\\
  $^2$ University of Chinese Academy of Sciences\\
  $^3$ École normale supérieure de Rennes, Inria, CNRS, IRISA\\
  $^4$ East China Normal University\\
  $^5$ Nanjing Institute of Software Technology, ISCAS\\
  $^6$ Guangdong Artificial Intelligence and Digital Economy Laboratory\\
  \texttt{lirj19, yangpf, xuebai, zhanglj@ios.ac.cn} \\
  \texttt{hanwei.zhang@irisa.fr} \\
  \texttt{chengchao@nj.iscas.ac.cn} \\
  \texttt{amzhou@cs.ecnu.edu.cn}
}

\begin{document}

\maketitle

\begin{abstract}


In this paper, we propose a framework of filter-based ensemble of deep neural networks (DNNs) to defend against adversarial attacks. The framework builds an ensemble of sub-models --- DNNs with differentiated preprocessing filters. From the theoretical perspective of DNN robustness, we argue that under the assumption of high quality of the filters, the weaker the correlations of the sensitivity of the filters are, the more robust the ensemble model tends to be, and this is corroborated by the experiments of transfer-based attacks. Correspondingly, we propose a principle that chooses the specific filters with smaller Pearson correlation coefficients, which ensures the diversity of the inputs received by DNNs, as well as the effectiveness of the entire framework against attacks. Our ensemble models are more robust than those constructed by previous defense methods like adversarial training, and even competitive with the classical ensemble of adversarial trained DNNs under adversarial attacks when the attacking radius is large.

\end{abstract}

\section{Introduction}

Recent research reveals that deliberately crafted adversarial perturbations succeed in leading Deep Neural Networks (DNNs) to make wrong predictions, not only when attackers are aware of the architecture of DNNs, i.e., white-box setting, but also when they only have access to the input-output pairs of DNNs, i.e., black-box setting.  This discovery exposes the potential danger in existing machine learning applications and encourages defenses against adversarial attacks.
These defenses are divided into two families, namely reactive defenses and proactive defenses. Reactive defenses~\cite{xu2017feature,guo2017countering,Sun_2019_CVPR,xie2017mitigating} aim to gain robustness by introducing an extra element to recognize or remove the adversarial context. Proactive defenses~\cite{goodfellow2014explaining,madry2017towards,lyu2015unified,chen2019improving} attempt to build networks inherently robust to adversarial attacks.

Transformations~\cite{xu2017feature,guo2017countering}, as a typical reactive approach, remove adversarial effects via applying simple filters. It is cheap but performs poorly against strong attacks, e.g., PGD~\cite{madry2017towards}, C\&W~\cite{carlini2017towards} and DeepFool~\cite{moosavi2016deepfool}. To augment the performance, randomness~\cite{raff2019barrage,prakash2018deflecting} and representation~\cite{moosavi2018divide,buckman2018thermometer,liu2019feature} are introduced into transformation. Transformation gains robustness but loses accuracy, since the original images are altered when it discards adversarial context. The networks learn from original data and cannot recognize the distorted information. 

Adversarial Training~\cite{goodfellow2014explaining,madry2017towards}, as a proactive defense, augments the training process with adversarial images such that the network learns the relative knowledge. To produce adversarial images, adversarial training employees a special attack. Since different attacks have different preferences, the model is vulnerable to unseen attacks. 
The ensemble is a solution to amend this drawback. It consists of several sub-models which learn from similar but different training sets, for instance, applying Gaussian noise to inputs and bootstrap~\cite{strauss2017ensemble} to augment robustness. Incorporating randomness stabilizes the performance of ensemble models. Random Self-Ensemble (RSE)~\cite{liu2018towards} adds random noise layers to prevent strong gradient-based attacks.  

Diversity is essential to ensembles. To increase the diversity of adversarial examples during the training, ensemble adversarial training~\cite{tramer2017ensemble} adds adversarial examples transferred from other pre-trained models. Ensemble-of-specialists~\cite{abbasi2017robustness} multiplies adversarial examples targeted over different incorrect labels. This defense is confirmed not robust enough~\cite{he2017adversarial}. Other than data augmentation, Adaptive Diversity Promoting (ADP)~\cite{pang2019improving} and Diversity Training~\cite{kariyappa2019improving} design a regularizer to encourage diversity. However, these methods either fail in defending against strong attacks or are too expensive because it needs too many sub-models to achieve a decent diversity. So the question arises: 
\begin{quote}
What is the advantageous diversity to improve the ensemble defenses against adversarial attacks?
\end{quote}

\paragraph{Contributions.} This work investigates the answer to this question. Inspired by the transformation defenses, we train sub-models with different front filters, such as dimension reduction, color quantization, and frequency filter. The model trained on a particular front filter is sensitive to a specific type of distortion. These front filters distort adversarial contexts. At the same time, training with transformed data allows models to learn and maintain accuracy on them.
We analyze the Pearson correlation coefficient among the models and the performance of the models and their ensemble. We infer that the sub-models with weakly correlated sensitivity constitute a more robust ensemble, and propose a simple and powerful defense framework for ensemble models based on the inference. Finally, the experimental results demonstrate that the proposed method improves the robust of the network against adversarial examples.

In the rest of the paper, we first introduce some basic notations and adversarial attacks in Section~\ref{sec:preliminary}. After elaborating the framework and the theoretical analysis in Section~\ref{sec:method}, we present experimental analysis and evaluation in Section~\ref{sec:experiment}. Finally, we conclude this paper in Section~\ref{sec:conclusion}.
\section{Preliminary}
\label{sec:preliminary}

\old{We recall some basic notions on deep neural networks and abstract interpretation.
For a vector $\bar x \in \mathbb{R}^n$, we use $x_i$ to denote its $i$-th entry.
For a matrix $W \in \mathbb R ^ {m \times n}$, $W_{i,j}$ denotes the entry in its $i$-th row and $j$-th column.}

\hw{In this section, we first state some basic notions of DNNs and the definition of DNN robustness in a local region.
Then we recall the norm-based robustness region, as well as the Lipschitz constant of DNNs. After that, we give a brief introduction to a few existing attacks used in our experiments.}

\newcommand{\proj}{\mathbb{P}}
\newcommand{\sign}{\operatorname{sign}}
\def \labell {\ell} 
 
\subsection{Deep Neural Network and Local Robustness}
\old{Our work concentrates on deep neural networks, which can be characterized as a function $ f :\mathbb{R}^m\to\mathbb{R}^n$. 
For classification tasks, a DNN usually chooses the output dimension with the largest score, i.e., $C_{ f }(\bm{x}):=\arg\max_{1\leq i\leq n}f_i(\bm{x})$, as its output label.}

\hw{Our work concentrates on the image classification task.
A DNN, which can be characterized as a function $ f :\mathbb{R}^m\to\mathbb{R}^n$, usually gives prediction by maximizing the output vector, i.e., $C_{ f }(\bm{x}):=\arg\max_{1\leq i\leq n}f_i(\bm{x})$, where $\bm{x} \in \mathbb{R}^m$ represents an image. To optimize the network, we minimize the cost function $\mathbb{J}( f ,\bm{x},\labell)$, in which $\labell$ is the ground truth.}




Intuitively, the local robustness of a DNN ensures the consistency of its behavior of a given input under certain perturbations, and a strict robustness condition ensures that there is no adversarial example around an input $\bm{x}$. Formally, the local robustness of a DNN can be defined as below.

\begin{definition}[DNN robustness]
\label{def:localrobustness}
Given a DNN $ f :\mathbb{R}^m \to \mathbb{R}^n$ and an input region $B\subset\mathbb{R}^m$, we say that $ f $ is (locally) \emph{robust} in $B$ if for any $\bm{x},\bm{x}'\in B$, we have $C_{ f }(\bm{x})=C_{ f }(\bm{x}')$.
\end{definition}

In a typical way, the region $B$ here is usually defined by the neighborhood of an input, where $L_p$-norm balls are commonly used. 
As the case of the $L_2$,
the neighborhood of an input $\bm{\bar{x}}$ bounded by the $L_2$-norm can be described as an $L_2$ ball: The $L_2$ (closed) ball with the center $\bm{\bar{x}} \in \mathbb R^n$ and the radius $r>0$ is defined as $B_2(\bm{\bar{x}},r)=\{\bm{x} \in \mathbb R^n \mid \|\bm{x}-\bm{\bar{x}}\|_2 \le r\}$.

\paragraph{Lipschitz constant of DNNs}
The Lipschitz constant of a function is a measure to indicates the maximum ratio between variations in the output space and variations in the input space.
In~\cite{RHK2018}, a DNN $f:\mathbb R^m \to \mathbb R^n$ is proved to be Lipschitz continuous. Namely, there exists $\mathcal{L} >0$, s.t. for any $ \bm{x}, \bm{x}' \in \mathbb R^m$, 
\[
\| f ( \bm{x})- f (\bm{x}')\|_2 \le \mathcal{L} \cdot \|\bm{x}-\bm{x}'\|_2,
\]
and here $\mathcal{L}$ is called a Lipschitz constant of $f$. Generally, DNNs with a smaller Lipschitz constant are likely to be more robust.


\subsection{Adversarial Attacks}

Adversarial attacking methods attempt to find an imperceptible perturbation leading to misclassification, also regarded as a testing method for network robustness. We present several fundamental untargeted attacks widely used in existing literature.  Hereafter, we denote a potential adversarial example as $\bm{x}'$, the adversarial perturbation as $\bm{\delta}$, and the gradient calculated from the cost function to input as $\nabla_{\bm{x}} \mathbb{J}(\bm{x})$.

\paragraph{Fast Gradient Sign Method (FGSM)}
Fast Gradient Sign Method (FGSM) by Goodfellow and his colleagues in 2014 ~\citep{goodfellow2014explaining} simply uses the one-step gradient to generate the perturbation:
\begin{equation}
	\bm{x}' = \bm{x} +   \bm{\delta} = \bm{x} - r \sign ( \nabla_{\bm{x}} \mathbb{J}(\bm{x}) ).
	\label{eq:fgsm}
\end{equation}

It is the perturbation that minimizes the first-order objective function for the constraint $\| \bm{\delta}\|_\infty = r$.

\old{\paragraph{Basic Iterative Method (BIM)}
It is possible to refine FGSM iteratively. BIM~\citep{kurakin2016physical} is the iterative version of FGSM. BIM initializes $\bm{x}'_0 := \bm{x}$ and then iterates by progressing in the opposite direction of the gradient with stepsize $\alpha$. The recurrence is therefore:
\begin{equation}
	\bm{x}'_{i+1} := \proj_{B_\infty(\bm{x}, r)}(\bm{x}'_i - \alpha \sign \nabla_{\bm{x}} \mathbb{J}(\bm{x}'_i)).
	\label{eq:ifgsm}
\end{equation}

Here, the ball $B_\infty(\bm{x}, r)$ of $L_\infty$-norm is centered in $\bm{x}$ with radius $r>\alpha$. Projection on the ball $B_\infty(\bm{x}, r)$ is defined as
\begin{align}
 \proj_{B_\infty(\bm{x}, r)}(\bm{x}') := \bm{x} + clip_{[-r,r]}(\bm{x}' - \bm{x}), \old{\textcolor{red}{What~does~'clip'~mean?}}
\end{align}
where $clip_{[-r,r]}(\cdot)$ denotes the clipping input into range $[-r,r]$ element-wise.
Thus, when the current solution remains inside the ball, the projection is not active. While the iterations calculate perturbations that get outside the ball, then the projection brings them back to its surface.}

\old{\paragraph{DeepFool}
DeepFool~\citep{moosavi2016deepfool} employs an objective function in order to modify the adversarial image into the most likely class. It seems easier to make $\bm{x}$ negative when the objective function has a positive but low value at $\bm{x}$.
In the first order for a targeted class, the minimum distortion necessary in $L_2$-norm is reached when $\bm{\delta}\propto - \nabla_{\bm{x}}\mathbb{J}(\bm{x}, \labell)$ with:
\begin{equation}
    \| \bm{\delta}\| = \frac{\mathbb{J}(\bm{x},\labell)}{\|\nabla_{\bm{x}}\mathbb{J}(\bm{x},\labell)\|}.
\end{equation}
It is best to target the class $\labell$ that will cause the addition of the smallest distortion.}

\paragraph{Projected Gradient Descent (PGD)}
PGD initializes $\bm{x}'_0 := \bm{x}$ and then iterates by progressing in the opposite direction of the gradient with stepsize $\alpha$. The accumulated distortion are projected onto an $L_p$-norm ball~\citep{madry2017towards}:
\begin{equation}
	\bm{x}'_{i+1} :=  \proj_{B_p(\bm{x}, r)}(\bm{x}'_i - \alpha \mathsf{n}(\nabla_{\bm{x}} \mathbb{J}(\bm{x}'_i))),
	\label{eq:pgd}
\end{equation}
where $\mathsf{n}(\bm{x}) := \bm{x} / \| \bm{x} \|_p$, and 
\begin{align}
 \proj_{B_p(\bm{x}, r)}(\bm{x}') := 
 \begin{cases}
 \bm{x}', & \text{if~}\| \bm{x}' - \bm{x} \|_p  < r,\\
 \bm{x} +  r \mathsf{n}({\bm{x}' - \bm{x}}), & \text{otherwise}.
 \end{cases}
\end{align}

The $L_p$-norm ball used for projection is $B_p(\bm{x}, r)$, centered in $\bm{x}$ with radius $r$. Again, the attack does not end when $\bm{x}'_i$ hits the boundary of the ball $B_p(\bm{x}, r)$ for the first time. It continues and seeks to minimize the objective function while remaining on the sphere.

\paragraph{Basic Iterative Method (BIM)}
BIM~\citep{kurakin2016physical}, as an iterative version of FGSM, it employees the sign of gradients of network iteratively with stepsize $\alpha$ to update adversarial perturbations. The principle of BIM is similar to a $L_\infty$-norm version of PGD. All the pixels in the adversarial example are clipped into range $[-r,r]$, i.e., adversarial perturbations are resized within the surface of an $L_\infty$-norm ball with radius $r$. The main difference is that PGD utilizes a random initialization and uses the gradients directly.

\paragraph{Backward Pass Differentiable Approximation (BPDA)}
BPDA~\citep{athalye2018obfuscated} allows attackers to generate adversarial perturbations targeted at the network with defenses as a whole. BPDA approximates derivatives by computing the forward pass normally and computing the backward pass using a differentiable approximation of the defense function. For instance, if it is impossible to calculate gradients through the transformation, BPDA generates adversarial examples by including the transformation during the forward pass and
replaces the transformation with an identity function during the backward pass under the assumption that the transformation output is close to the original input. 

\section{Methodology}
\label{sec:method}
In this section, we state the structure of our ensemble framework. A filter is an image transformation that extracts some important features of the original image. 
We embed a filter in each sub-model as the core component,
which provides the diversity of sub-models. Then we analyze the relationship between the correlations of the filters and the local robustness, which induces a principle of choosing the optimal filter combination. This improves the ensemble defense against adversarial attacks.

\subsection{Filter-based Ensemble}
\begin{figure}[ht]
    \centering
        \centering
        \includegraphics[width=0.95\linewidth]{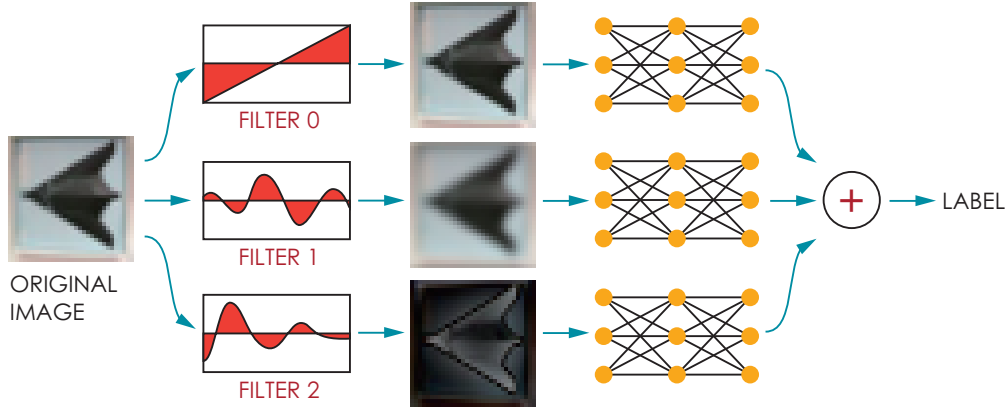}
    \caption{An illustration of the structure of our ensemble model.}
    \label{fig:stucture}
\end{figure}

\old{The structure of our ensemble model is shown in Figure~\ref{fig:stucture}.}

The structure of an ensemble model is shown in Figure~\ref{fig:stucture}. The model consists of several different sub-models, each of them equipped with a DNN and a front filter.
\hw{As shown in Figure~\ref{fig:stucture},}
an input image is pre-processed by three
different filters, and the obtained results are then respectively fed into three DNNs, which classify them and output the classification label individually. In the end, the results of each sub-models are combined by a voting mechanism.
\hw{Formally, a certain sub-model applies a front filter, denoted by $\zeta(\bm{x}):\R^m \to \R^{s}$, on the original inputs, and the DNN model follows the filter by $ f (\bm{x}):\R^{s} \to \R^n$.}
Then an ensemble model with $k$ sub-models $(\zeta_i,f_i)_{i=1}^k$ can be expressed as
\[
\mathcal{E}(\bm{x})= \mathrm{Vote} ( \{C_{ f_i } \circ \zeta_i (\bm{x})\}_{i=1}^k )\;,
\]
where the function $\mathrm{Vote}$ outputs the mode of the results of the sub-models, i.e., the classification label which appears the most times.

We call a input $\bm{x}$ stable for an ensemble model $\mathcal{E}$ if the output labels of all the sub-models in $\mathcal{E}$ are consistent. It is easy to obtain the following proposition.
\begin{proposition}\label{prop:attboth}
The ensemble model $\mathcal{E}$ cannot be attacked at a stable input $\bm{x}$ by the perturbation $\bm{\delta}$ if 
any two sub-models are not attacked simultaneously at their respective inputs $\zeta(\bm{x})$ by the perturbations $\zeta(\bm{x}+\bm{\delta})-\zeta(\bm{x})$.
\end{proposition}

\old{To defend against the adversarial attack, we hope to build a more robust ensemble model. Unlike a lot of recent defense approaches, we don not focus on the training skills for the sub-models to gain a good defense effect. 
On the contrary, our main idea is enhance the diversity of the sub-models --- using differentiated front filters to pass the partial feature of the inputs, which makes the perturbation generated by adversarial attacks cannot affect every DNN ensembled.}

\hw{To defend against an adversarial attack, we propose to build a more robust ensemble model.
Unlike network-based defenses like adversarial training, we do not focus on training skills for the sub-models to improve robustness. 
On the contrary, our principle is to enhance the diversity of sub-models by extracting partial features of inputs using differentiated front filters. Since it is hard for an adversarial attack to effectively affect all sub-models at the same time, the ensemble model achieves better robustness from the diversity of front filters.}

So, the key to establish this ensemble model is how to choose a proper filter combination that provide both accuracy and robustness.
In the following, we explain how we gain a more robust ensemble model through the relation among the filters.

\subsection{Low Correlation Implies Strong Robustness} 
\label{Sect:Correlation}
In this subsection, we give a theoretical description on the intuition that a low correlation of the sensitivity of two filters implies a more robust ensemble model under the assumption that the filters are of high quality, and this will guide us to choose the optimal filter combination from the candidates.

For an input $\bm{x}\in\R^m$ and a perturbation $\bm{\delta}\in\R^m$, 
we define a function $r:\R^m \times \R^m \to \R_{\ge 0}$ as
\begin{equation}\label{eq:rsensi}
    r(\bm{x},\bm{\delta})=\|\zeta(\bm{x}+\bm{\delta})-\zeta(\bm{x})\|_2
\end{equation}
to measure the sensitivity of a filter $\zeta$, i.e., the $L_2$-norm of the
perturbation affecting the input of the DNN in the sub-model.
Considering $r$ as a random variable,
we invoke the \emph{Pearson correlation coefficient} 
to evaluate the correlation of the sensitivity of two filters,
which is expressed as
\begin{equation}\label{eq:corre}
    \rho_{r_1,r_2}=\frac{\E[r_1r_2]-\E[r_1]\E[r_2]}{\sqrt{\mathrm{Var}[r_1]\mathrm{Var}[r_2]}}\;.
\end{equation}
We assume that the filters in the ensemble model are of high quality, which means that the interpretation of the difference of two images by each filter sincerely reflects their semantics difference in statistics, i.e.,  the random variables $r_1$ and $r_2$ are identically distributed. Under this assumption,
the equation~\eqref{eq:corre} indicates that $\E[r_1r_2]$ is monotonically increasing w.r.t. $\rho_{r_1,r_2}$. 

For a certain DNN $ f $ and an input $\bm{x}$ classified into label $\labell$, we define the score difference by $\Delta_{ f }(\bm{x})= \min_{i \neq \labell}\left( f _\labell(\bm{x}) - 
 f _i(\bm{x})\right)$.
Then the robust radius at an input $\bm{\bar{x}}$ can be 
estimated according to the following lemma.
\begin{lemma}[\cite{faoc}]
Consider a DNN defined by $ f (\bm{x}):\R^m\to\R^n$, whose Lipschitz constant
is $\mathcal{L}_{ f }$. Then for an input $\bm{\bar{x}}$, the DNN is robust
in $B_2(\bm{\bar{x}}, r)$ with $r < \tfrac{\Delta_{ f }(\bm{\bar{x}})}{\sqrt{2}\mathcal{L}_{ f }}$.
\end{lemma}


Then, we can infer that
two DNNs $ f _1$ and $ f _2$ cannot be attacked simultaneously at $\bm{\bar{x}}_1$ and $\bm{\bar{x}}_2$ by the perturbations $\bm{\delta}_1$ and $\bm{\delta}_2$ respectively,
if
\begin{equation}\label{eq:RobProd}
\|\bm{\delta}_1\|_2\|\bm{\delta}_2\|_2 < \frac{\Delta_{ f _1}(\bm{\bar{x}}_1) \Delta_{ f _2}(\bm{\bar{x}}_2)}
{2\mathcal{L}_{ f _1}\mathcal{L}_{ f _2}}\;.
\end{equation}
In our framework, $\bm{\bar{x}}_1$ and $\bm{\bar{x}}_2$ are the input processed by two filters, and $\|\bm{\delta}_1\|_2$ and $\|\bm{\delta}_2\|_2$ are the sensitivity calculated by \eqref{eq:rsensi}. 
 It is clear that the right part in the inequality~\eqref{eq:RobProd} is determined by the structure and parameters of the DNNs, while the left part is determined by the front filters. 
 
The expectation $\mathrm E[r_1,r_2]$ is a statistical description of the item $\|\bm{\delta}_1\|_2\|\bm{\delta}_2\|_2$ in \eqref{eq:RobProd}: A small expectation $\mathrm E[r_1,r_2]$ imples that the value of $\|\bm{\delta}_1\|_2\|\bm{\delta}_2\|_2$ tends to be small statistically.
Consequently, by combining \eqref{eq:corre}, \eqref{eq:RobProd} and Proposition~\ref{prop:attboth},
\old{we infer that the filters with independent sensitivity
implies a more robust ensemble model.}
we infer that low correlation of the sensitivity among filters
implies strong robustness of ensemble models.
This leads to our principle as `{\bf minimum correlation coefficients}' for choosing filter combinations, i.e., to optimize the robustness of our ensemble model, we choose the filters among which the correlation is the weakest.

\begin{remark}
Note that the cosine similarity of $\zeta(\bm{x}+\bm{\labell})-\zeta(\bm{x})$ between two filters
is also a measure of their correlation.
However, we only consider the sensitivity from the perspective of magnitudes of perturbation vectors generated by filters, because the gradient of the entire sub-model depends on both the filter and the DNN, so we do not choose to analyze their directions without considering the following DNNs.
\end{remark}


\subsection{Filter Candidates} The original image is prop-processed by a filter before it is sent to the network. Therefore, some information is discarded and thus the overall entropy of the image 
is reduced. It is also regarded as a manual feature extraction procedure that extracts the most important features that benefit the task. It is  generally harder to attack the filtered image  because there is less information that the attacking methods can utilize. The filters we use are categorized into the following four classes.

\paragraph{Dimension Reduction} 
The easiest way to reduce the entropy of an image is to reduce its dimensionality. Color pictures have three dimensions, i.e., length, width, and color channels. It is simple to reduce the first two dimensions by downsizing an image. Grayscale transformation can compress the color channels into one grayscale channel. Generally, downsizing and grayscale transformation preserve the overview of the original image with certain loss of details. Bilinear interpolation is used in downsizing filters. We use the ITU-R BT.601{\footnote{BT.601 : Studio encoding parameters of digital television for standard 4:3 and wide screen 16:9 aspect ratios: \url{https://www.itu.int/rec/R-REC-BT.601-7-201103-I/en}}} luma transformation for the grayscale filter.



\paragraph{Color Quantization} Another way to reduce the complexity of an image is to reduce its number of colors. The size of a CIFAR-10 image is $32\times32$. It may have $1\,024$ colors at most. But it can still be recognizable using much fewer colors. 
The fast octree algorithm~\cite{fastoctree} is adopted to reduce the colors of the image. A full-size octree with a depth of seven can be used to partition the RGB color space. The octree subdivides the colorspace into eight octants recursively. Each leaf node of the octree represents an individual color. The fast octree algorithm builds the tree according to the color of a given image and merges the leaf nodes when the number of colors overflows.

\paragraph{Frequency Filters} In digital image processing, frequency filters are commonly applied to extract useful features from pictures. The high-frequency features are usually the noise and the details of the original image, and the low-frequency features are often its overview. The high-pass filters suppress the low-frequency features, and the low-pass filters do the opposite.
Our low-pass and high-pass filters are based on the discrete Fourier Transform. 
Via shifting the low-frequency part to the center of the spectrum and multiplying it by a Gaussian mask (high-pass mask) element-wise, we obtain the low-pass (high-pass) filtered image. 

\paragraph{Data Discretization} 
Inputs for a DNN can be any real value, while the 8-bit RGB color model takes integer values in the range of $[0,255]$. To keep the practical meaning of an input, real numbers are approximated by its closest integer. This is essential since the DNN for image classification should have actual image data instead of arbitrary inputs. When we use iterative methods to attack the DNN, discretization can help generate practical adversarial examples. In our ensemble model, every sub-model trained with the original data is equipped with a discretization filter.
 

\section{Experimental Evaluation}
\label{sec:experiment}

\old{In this section, the experimental results are provided to demonstrate the performance of our method against adversarial perturbations. We evaluate the correlation between filters stated in Sect.~\ref{Sect:Correlation}. The adversarial examples of the network trained with original data are generated and tested on the network trained with filtered data to measure the transferability empirically. We choose two filters accordingly for the ensemble model and compare the adversarial robustness with the previous ensemble defense and the adversarial training.}

\hw{In this section, we demonstrate the experimental results to support our inference and our method. We first give a brief introduction of the experimental settings, and then show our study of the correlation between filters in Section~\ref{Sect:Correlation}. We measure the robustness of our models with respect to transferability, via calculating the accuracy of our models on adversarial examples produced by attacking the original network. In the end, two front filters are chosen by `minimum correlation coefficients' to constitute the ensemble defense. We compare its robustness with adversarial training. We generate adversarial examples under various attacks implemented in FoolBox~\cite{rauber2017foolbox,rauber2017foolboxnative}. The FoolBox version is 3.31, and the license is MIT license. All experiments are conducted on a Windows 10 laptop with Intel i7-9750H, GTX 2060, and 16G RAM.}

\paragraph{Dataset and Network} We train our models on the CIFAR-10~\cite{krizhevsky2009learning} dataset under the MIT license. The CIFAR-10 dataset contains $60\,000$ RGB images in total with ten exclusive classes. We train our ResNet18~\cite{he2016deep} models on the $50\,000$ training images and test them on the $10\,000$ testing images. We use the stochastic gradient descent optimizer for training with $0.1$, $0.01$, and $0.001$ as the learning rate successively.

\subsection{The Optimal Filter Combination for Ensemble}
\old{Our principle for choosing filters is to pick the least correlated filters with the minimum correlation coefficients. We first calculate the correlation coefficients for each pair of the candidate filters and then evaluate the sub-models trained with different filters using the transfer-based attack.}

\hw{We analyze the Pearson correlation coefficients for each pair of the candidate filters and pick the least correlated filters for the ensemble with the minimum correlation coefficients. We also evaluate the robustness of sub-models on adversarial examples produced on the original network.}

\begin{figure}[htbp]
\centering
\includegraphics[width=0.5\textwidth]{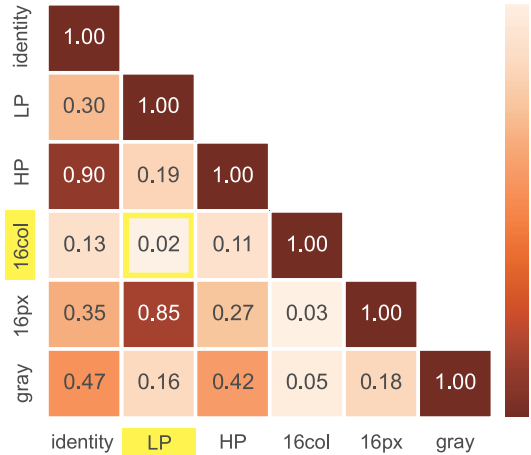}
\caption{The Pearson correlation coefficient between the sensitivity of different filtered data. The low-pass and the $16$ color reduction filter is marked yellow since they have the smallest coefficient.}
\label{fig:correlation}
\end{figure}

\old{\paragraph{Statistical Correlation Analysis for Filters} We apply noise of size $\epsilon \leq 20/255$ to 100 images randomly picked from the test set and measure the correlation using the mentioned Pearson correlation coefficient~\eqref{eq:corre}. The result is shown in Figure~\ref{fig:correlation}. The high-pass filtered data is most correlated to the original data, and the grayscale filtered is the second. It infers that the adversarial example of the network trained with the original data may be easy to transfer to the network trained with these filtered data. The downsizing filter is strongly correlated to the low-pass filter, implying that it is probable to attack these two filters together. It is interesting to see that the 16 colour reduction filter shows little correlation to the low-pass filter, and they have a relatively low correlation to the original data. It gives us the confidence to use the low-pass and the 16 colour reduction filter with the original data to build a robust ensemble model.}

\hw{\paragraph{Statistical Correlation Analysis for Filters} We apply noise of size $\epsilon \leq 20/255$ to $100$ images randomly picked from the test set and evaluate Pearson correlation coefficient according to~\eqref{eq:corre}. 
As reported by Figure~\ref{fig:correlation}, the correlation coefficient between the high-pass filtered data and the original inputs is the largest, i.e., $0.90$. The grayscale filtered data gets $0.47$, the second to the original inputs. It indicates that adversarial examples produced on the original network easily transfer to the sub-models with these filters. The downsizing filter 
strongly correlates With the low-pass filter, implying that they are probably deceived by the same adversarial examples.
The $16$ color reduction filter shows little correlation, i.e., $0.02$, to the low-pass filter, and they both have a relatively low correlation to the original data, i.e., $0.13$ and $0.30$. According to minimum correlation coefficients, the robust ensemble model includes the original network and the two sub-models trained with the low-pass filter and the $16$ color reduction filter.}


\old{\paragraph{Transfer-based Attack Analysis} We generate adversarial examples of the original network under the FGSM and the PGD attack with different distances and test them on the network trained with filtered data. The result is in Fig.~\ref{fig:transfAtt}. The outcome shows that the low-pass, colour quantization, and downsizing filters have a comparably better defence against the transfer-based attack. On the contrary, the grayscale and the high-pass filters are vulnerable to the transfer-based attack. So the corresponding two filters are not competitive to other filters since we want to build a robust ensemble model. The result of the transfer-based attack agrees with the previous correlation analysis.}

\hw{\paragraph{Transfer-based Attack Analysis} We generate adversarial examples against the original network by FGSM and PGD attacks with different values of attacking radius $\epsilon$ and test them on the sub-models trained with filtered data. In Figure~\ref{fig:transfAtt}, the sub-models with the low-pass, color quantization, and downsizing filters perform better than the original network against both FGSM and PGD attacks. The accuracy of these sub-models remains above $50\%$ when the attacking radius is $20/255$ under the FGSM attack. The PGD attack is more powerful against the original network and drops its accuracy to nearly $0$. The sub-model with a downsizing filter has the lowest accuracy among the three sub-models, which is $62.39\%$ under the PGD attack. However, the sub-models with the grayscale and the high-pass filters are vulnerable to the transfer-based attack. These sub-models have lower accuracy than the original network under the FGSM attack with $\epsilon \geq 10/255$. It is consistent with our analysis in Section~\ref{Sect:Correlation}, which suggests that these two filters should not be part of the ensemble. }

\old{
\begin{figure}[htbp]
    \centering
    \subfloat{ 
        \includegraphics[width=0.46\linewidth]{fig/fig_FGSM.eps}}
    \qquad
    \subfloat{ 
        \includegraphics[width=0.46\linewidth]{fig/fig_PGD.eps}}
    \caption{Transfer Attck by FGSM (left) and PGD (right) }
   \label{fig:transfAtt}
\end{figure}
}

\hw{
\begin{figure}[htbp]
\centering
\begin{tabular}{c c}
    \includegraphics[width=0.46\linewidth]{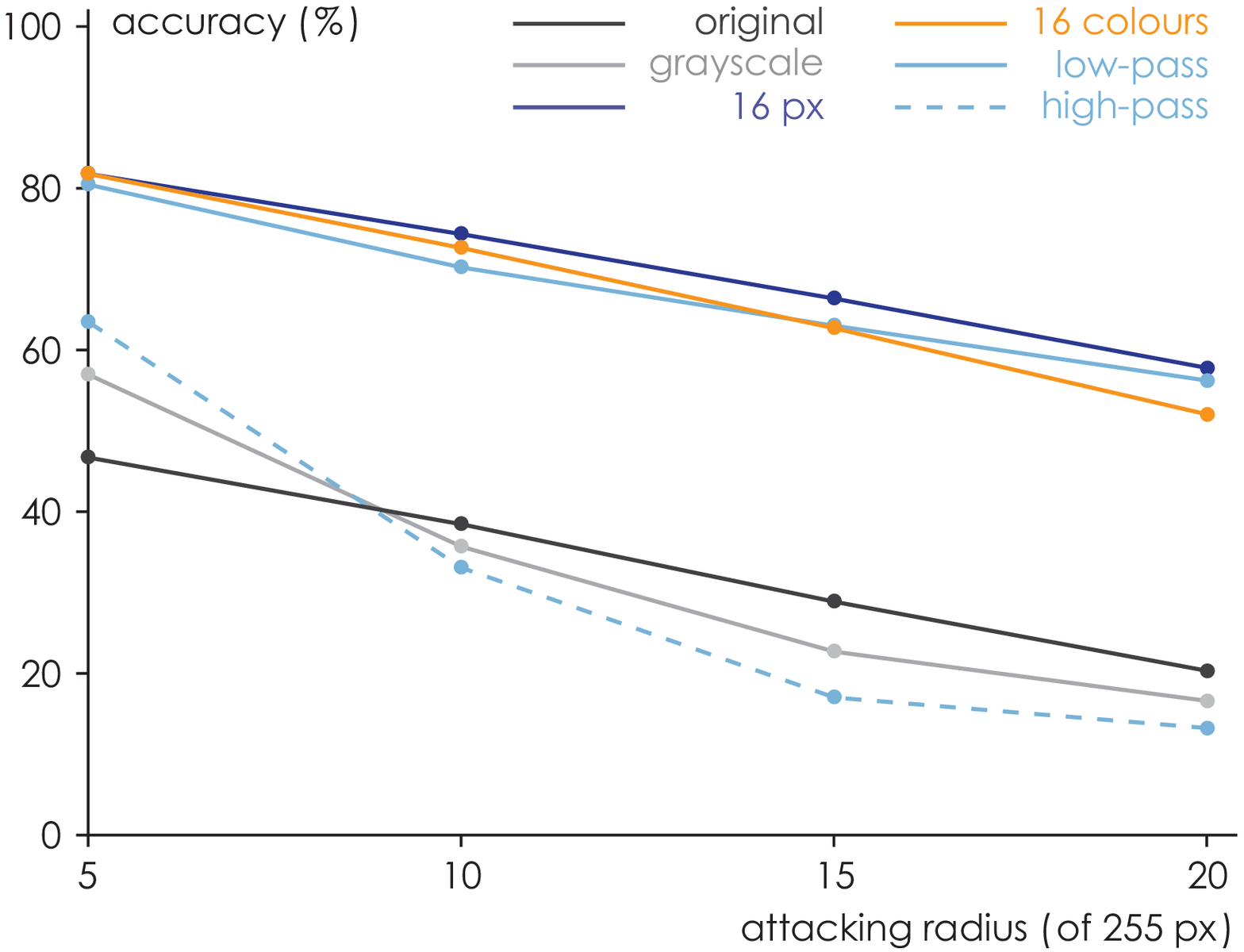} & 
    \includegraphics[width=0.46\linewidth]{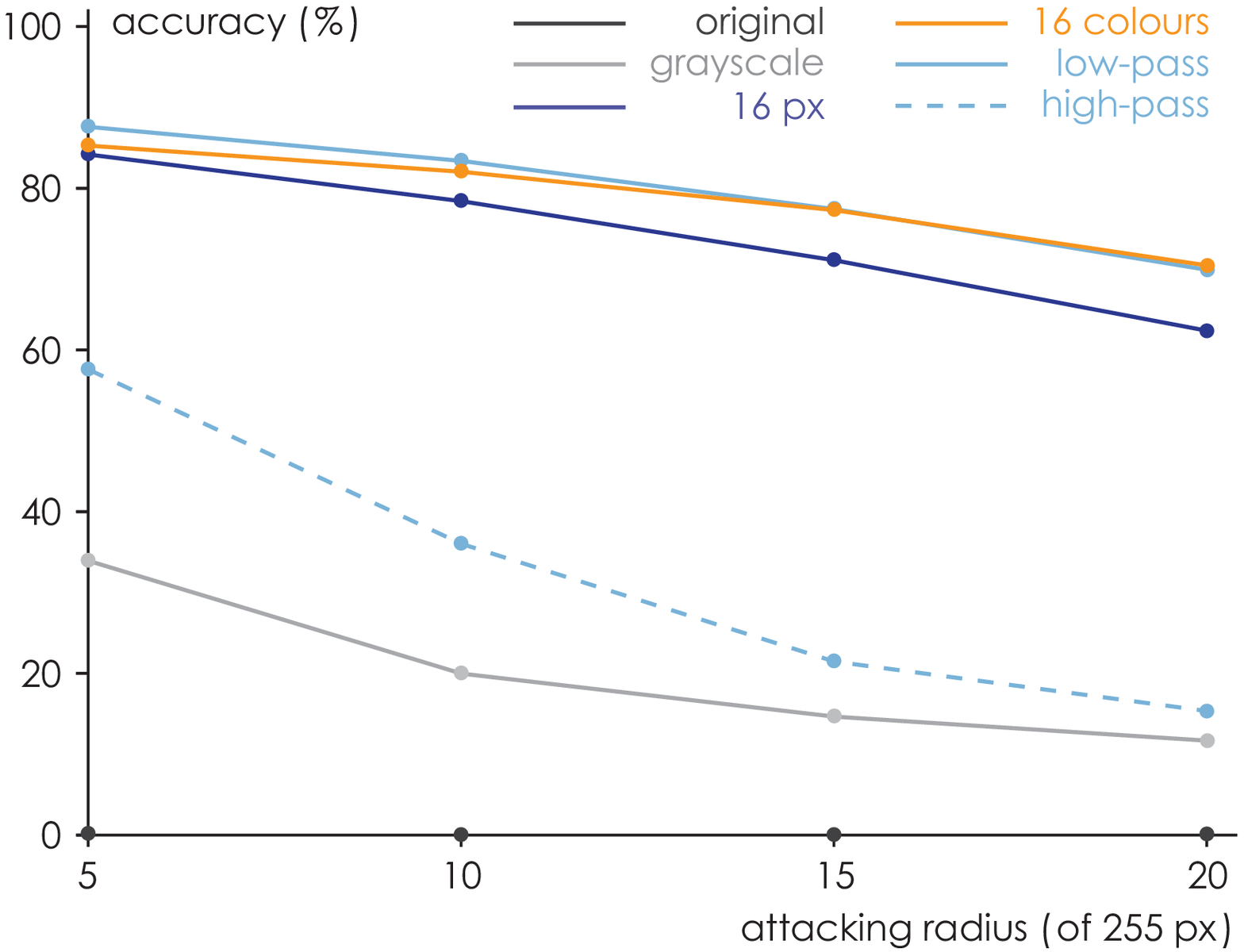}\\
    (a) FGSM & (b) PGD
\end{tabular}    
\caption{The transferablity of the adversarial examples against the original network generated by FGSM (left) and PGD (right). Note that the accuracy of the original network is near $0$ under the PGD attack.}
\label{fig:transfAtt}
\end{figure}
}

\old{\subsection{Comparison with Different Ensemble Methods}
According to the results of the correlation analysis and the transfer-based attack, we choose the 16 colour reduction and the low-pass filter for the ensemble model.  We also include a model trained with original data for the ensemble to maintain state-of-the-art accuracy on clean data. In this section, we compare the adversarial robustness of our ensemble model with the previous ensemble defence in \cite{ensembledefense} and the worst-case ensemble in our setting. We choose the BPDA attack based on the BIM attack to attack the ensemble model with filters. The number of iterations is set to be $20$, and the step size is equal to $\epsilon/10$. The result is shown in Fig.~\ref{fig:otherensemble}.}

\hw{\subsection{Comparison with Different Ensemble Methods}
In this section, we compare the adversarial accuracy of our ensemble model with different ensemble models. The details of each ensemble model are as follows:
\begin{itemize}
    \item \textbf{Minimum correlated Ensemble} According to statistical correlation analysis for filters, we choose the $16$ color reduction filter and the low-pass filter for the ensemble, which have the lowest correlation. The ensemble includes the original network to maintain state-of-the-art accuracy on clean data.
    \item \textbf{Maximum correlated Ensemble} The worst-case situation suggested by statistical correlation analysis is to constitute the ensemble with the original network and the two sub-models with the high-pass and grayscale filters, whose correlation is the highest as shown in Figure~\ref{fig:correlation}.
    \item \textbf{Gaussian Noise Ensemble~\cite{strauss2017ensemble}} The ensemble of models trained with Gaussian noise is the most robust model compared to the other ensemble methods~\cite{strauss2017ensemble}. We train three sub-models with Gaussian noise to build the ensemble model.
\end{itemize}
We compare the minimum correlated model to the Gaussian noise ensemble model and the maximum correlated ensemble model. We choose the BPDA attack based on the BIM to attack ensemble models. We use the sum of the gradient of sub-models to attack the ensemble model as a whole. The number of iterations is $20$, and the step size is $\epsilon/10$. Hereafter, the vote-based ensemble follows the voting mechanism described in Section~\ref{sec:method}, and the score-based ensemble outputs the class with the maximum average score.}


\hw{According to Figure~\ref{fig:otherensemble}(a), when the disturbance is $5/255$, $10/255$, $15/255$, and $20/255$, the score-based accuracy of the minimum correlated ensemble model is $27.52\%$, $22.4\%$, $14.33\%$, and $8.59\%$ higher than the Gaussian noise model, and is $25.54\%$, $15.46\%$, $9.53\%$, and $5.59\%$ higher than the maximum correlated model, respectively.
The minimum correlated ensemble model also has higher vote-based adversarial accuracy than the Gaussian ensemble model and the maximum correlated ensemble model.
It agrees with the previous analysis that the ensemble model with less correlated sub-models obtains better adversarial robustness.  }


\old{Figure~\ref{fig:otherensemble}(b) demonstrates the accuracy of sub-models when attacking the ensemble model as a whole. When we attack the ensemble model trained with Gaussian noise, the accuracy of its three sub-models decreases in a similar pattern. However, the sub-models of our ensemble model perform differently. The accuracy of the sub-model with the low-pass filter stabilizes when the radius is larger than $5$. The accuracy of the sub-model with the 16 color filter decreases but still above the curves of sub-models of Gaussian ensemble. The original network performs the worst. This result justifies that our ensemble model improves robustness against adversarial attacks by introducing advantageous diversity.}

\hw{Figure~\ref{fig:otherensemble}(b) demonstrates the accuracy of sub-models when the ensemble model is attacked  as a whole. When we attack the Gaussian noise ensemble model, the accuracy differences among its sub-models are close, which is $6.09\%$ at most. Also, the accuracy of its three sub-models decreases in a similar pattern. However, the sub-models of the minimum correlated ensemble model perform differently. The accuracy of the sub-model with the low-pass filter stabilizes when the radius is larger than $5/255$. The accuracy of the sub-model with the 16 color filter and the original network decreases with a relatively larger accuracy difference of $8.48\%$ at least. It justifies that our minimum correlated ensemble model improves robustness against adversarial attacks by introducing advantageous diversity.}

\old{
\begin{figure}[t]
    \centering
    \subfloat{ 
        \includegraphics[width=0.46\linewidth]{fig/fig_vsGauss.eps}}
    \qquad
    \subfloat{ 
        \includegraphics[width=0.46\linewidth]{fig/fig_sixlinesGauss.eps}}
    \caption{Comparison with Gauss Ensembling.}
   \label{fig:otherensemble}
\end{figure}
}

\hw{
\begin{figure}[htbp]
\centering
\begin{tabular}{c c}
    \includegraphics[width=0.46\linewidth]{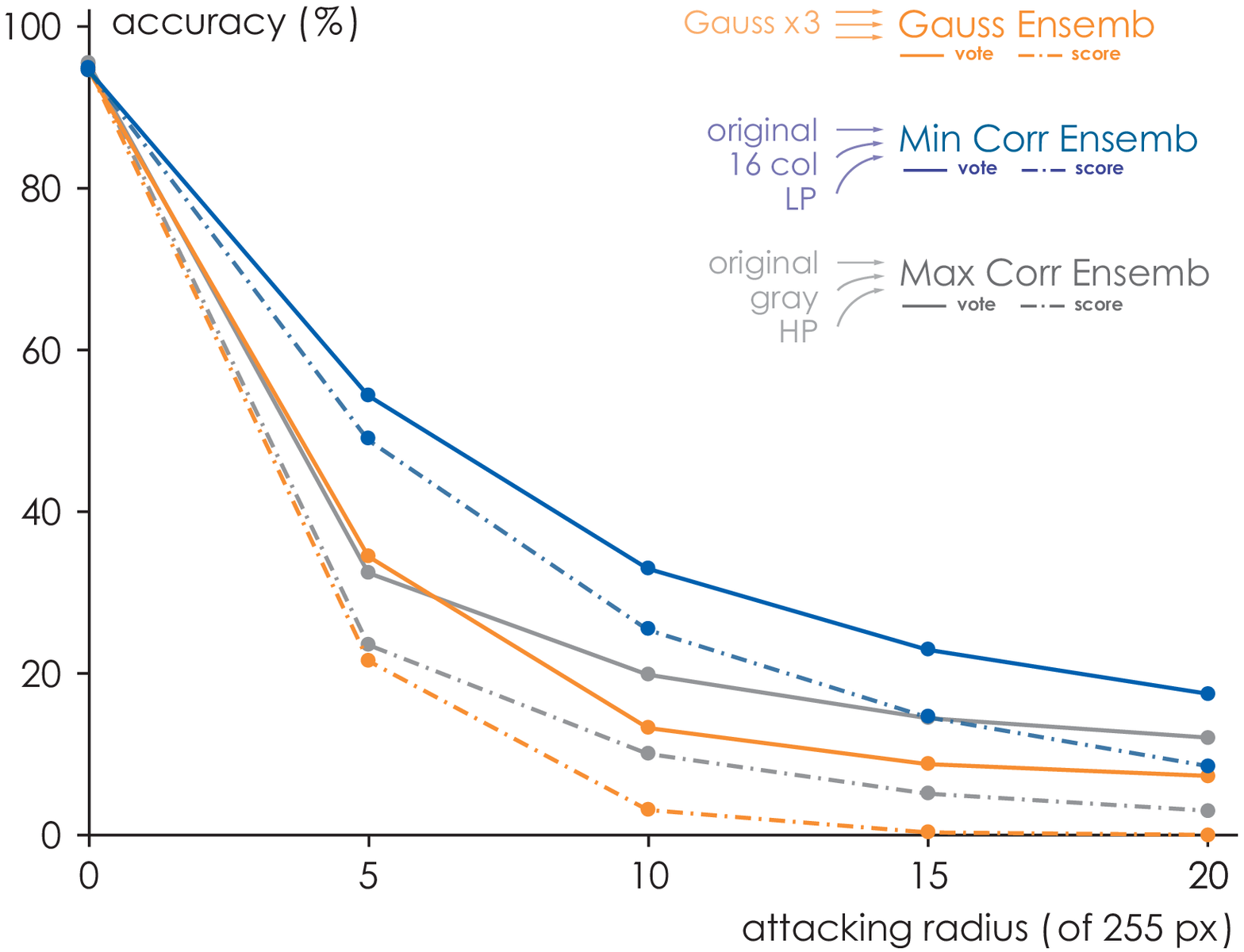} & 
    \includegraphics[width=0.46\linewidth]{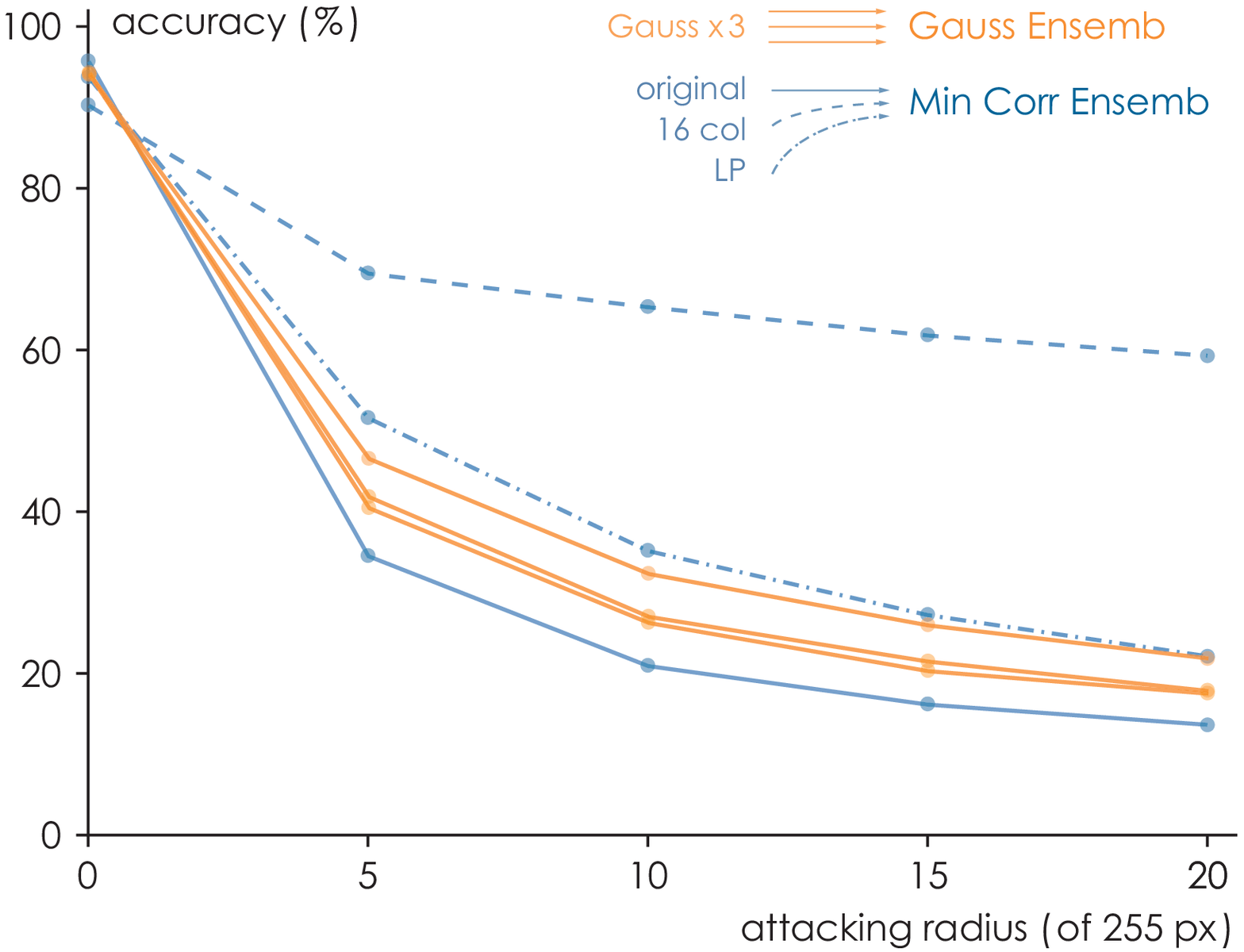}\\
    (a) & (b) 
\end{tabular}    
\caption{Adversarial accuracy of minimum correlated ensemble model,
the maximum correlated ensemble model and the Gaussian noise ensemble model under the BPDA attack. The sub-model accuracy of the minimum correlated ensemble model and the Gaussian noise ensemble model is presented on the right.}
\label{fig:otherensemble}
\end{figure}
}

\subsection{Comparison with Adversarial Training}
Adversarial training is one of the most effective methods to improve the robustness of a DNN. We use the method proposed in \cite{shafahi2019adversarial} and compare the robustness of our minimum correlated ensemble model with the adversarial training. The adversarial training procedure takes $4$ iterations with the maximal perturbation size $\epsilon=8/255$.

\old{\paragraph{Comparison with a single Adversarial Training Model}
We first compare our ensemble model with one single adversarial training model. The grey line in the left part of Fig.\ref{fig:ATcompare} shows the adversarial robustness of a single adversarially trained model. It is interesting to see that our ensemble model has better adversarial accuracy under all perturbations. Note that every sub-model in our ensemble model has no defence mechanisms acting on the network. In other words, our ensemble-based defence can build robust models competing with adversarial training without manipulating the network.
}
\hw{\paragraph{Comparison with a single Adversarial Training Model}
We first compare our ensemble model with one single adversarial training model. The grey line in Figure~\ref{fig:ATcompare}(a) shows the adversarial robustness of a single adversarially trained model. Our ensemble model has better adversarial accuracy under all perturbations. The score-based accuracy of our method is $12.53\%$ higher than the single adversarial trained model at $\epsilon=10$. It is worth highlighting that every sub-model in our ensemble model has no defense mechanisms acting on the network. In other words, our ensemble-based defense can build robust models competing with adversarial training without manipulating the network.}

\paragraph{Comparison with Adversarial Training Ensembles}
We compare our method with the ensemble of three independent adversarially trained models. The orange lines in Figure~\ref{fig:ATcompare}(a) show the performance of the ensemble model using adversarial training. Remarkably, our score-based ensemble model has better accuracy than its counterpart with adversarial training when the perturbation size is large, i.e., $\epsilon \geq 10/255$. Meanwhile, our vote-based ensemble is very close to the one with adversarial training. Our ensemble model has a comparable defense to the ensemble of adversarially trained sub-models.

The orange lines in the right part of Figure~\ref{fig:ATcompare}(b) depict the accuracy of adversarially trained sub-models when the ensemble model is attacked as a whole. Comparing with the sub-models trained with Gaussian noise in Figure~\ref{fig:otherensemble}, adversarial training does not significantly improve the diversity between sub-models. The accuracy difference between sub-models is still relatively small, and that means the attacking methods can affect different sub-models simultaneously.

Since the adversarial training works on the network level and our method works on the data level, it is natural to combine these two methods. We demonstrate our ensemble model with adversarial training in Figure~\ref{fig:ATcompare}(a) using the cyan lines. Our ensemble model with adversarial training reaches a better robustness performance in both score-based and vote-based settings. The vote-based ensemble model achieves $71.53\%$ accuracy at $\epsilon = 5/255$. Conclusively, we build an ensemble model with high adversarial robustness using both our filter-based defense and adversarial training.

\old{
\begin{figure}[t]
    \centering
    \subfloat{ 
        \includegraphics[width=0.46\linewidth]{fig/fig_vsESMAT.eps}}
    \qquad
    \subfloat{ 
        \includegraphics[width=0.46\linewidth]{fig/fig_sixlines.eps}}
    \caption{Comparison with Adversarial Training.}
   \label{fig:ATcompare}
\end{figure}
}

\hw{
\begin{figure}[htbp]
\centering
\begin{tabular}{c c}
    \includegraphics[width=0.46\linewidth]{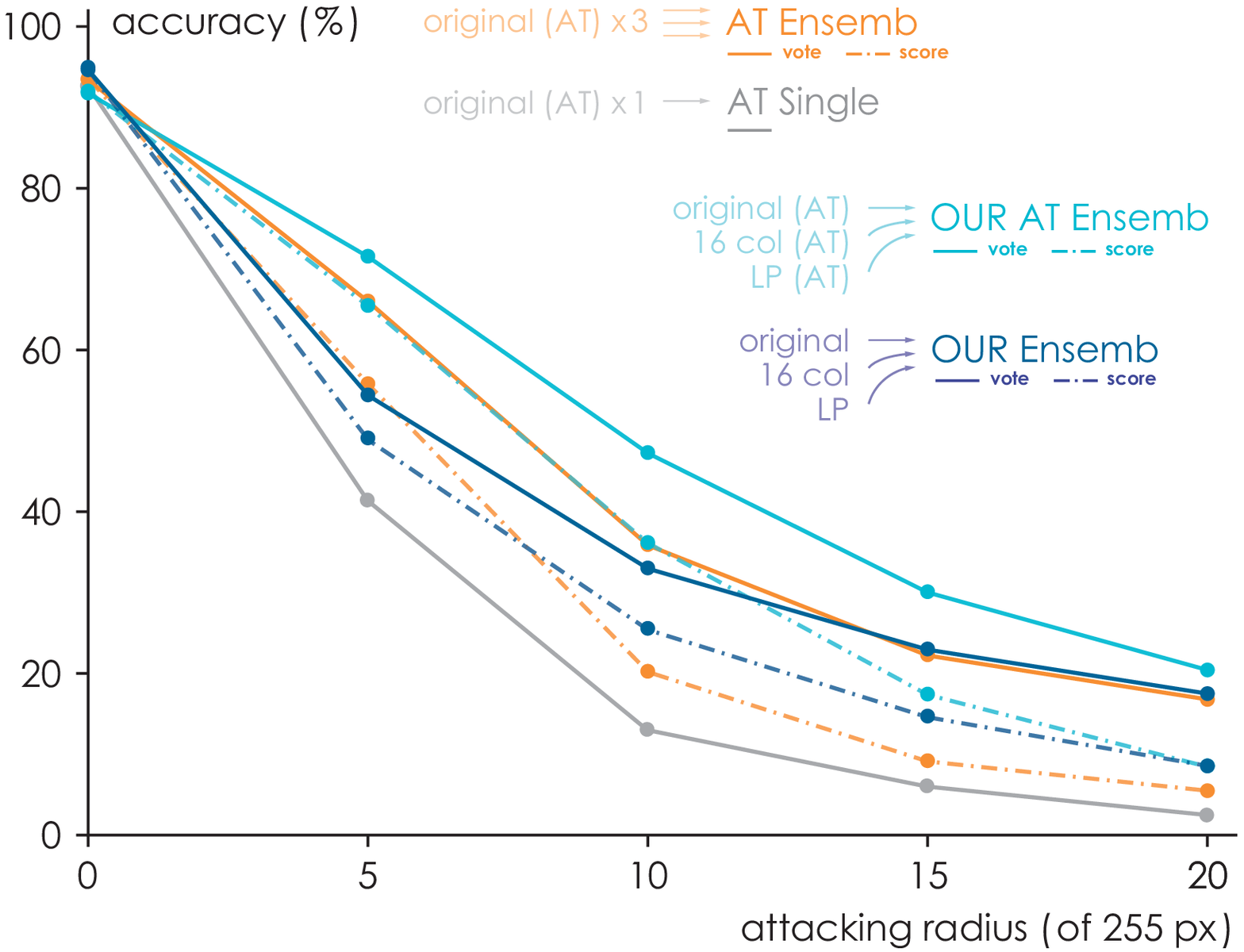} & 
    \includegraphics[width=0.46\linewidth]{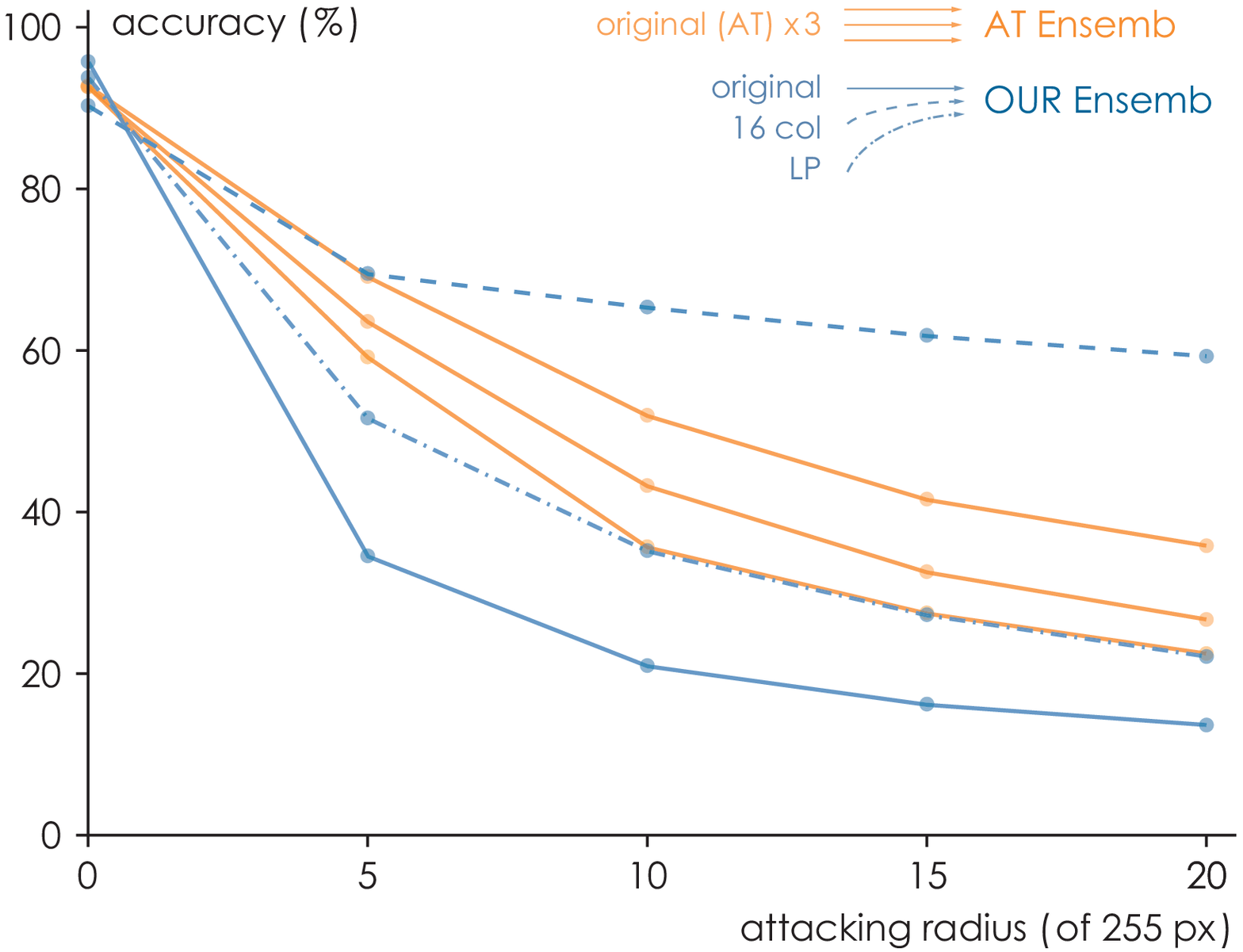}\\
    (a) & (b) 
\end{tabular}    
\caption{Adversarial accuracy of the single adversarial trained (AT) model, ensemble model with adversarial training, our ensemble model with and without adversarial training under the BPDA attack. The sub-model accuracy of our ensemble model and the ensemble model with adversarial training is presented on the right.}
\label{fig:ATcompare}
\end{figure}
}

\section{Conclusion}
\label{sec:conclusion}

In this work, we investigate the advantageous diversity of the ensemble model against adversarial attacks. By studying the robustness of ensemble DNNs and the Pearson correlation coefficient among models trained with filters, we propose the "minimum correlation coefficients" principle for choosing filters, which is instrumental in building the ensemble defense. 

Beyond existing ensemble defenses, we consider the diversity of ensemble models with a new perspective. We obtain the diversity from the filtered training data and confirm it  experimentally. We observe that our ensemble model without adversarial information is more robust against adversarial attacks than adversarial training models.

Our discovery not only contributes to proposing a decent robust ensemble model but also supplies data diversity. As our future work, it is interesting to study further how much robustness we could gain from data diversity and model ensemble. We are also considering extending our framework to larger datasets like ImageNet and training our sub-models using different network structures.

\medskip
\bibliography{paper}
\bibliographystyle{plain}

\end{document}